\documentclass[journal,twoside]{IEEEtran}
\usepackage{cite}
\usepackage{amsmath,amssymb,amsfonts}
\usepackage{graphicx}
\graphicspath{{figures/}}
\usepackage{algorithm,algorithmic}
\usepackage{textcomp}
\usepackage{booktabs}
\usepackage{makecell}
\usepackage{stfloats}
\usepackage[hidelinks]{hyperref}
\usepackage[switch]{lineno} 

\def\BibTeX{{\rm B\kern-.05em{\sc i\kern-.025em b}\kern-.08em T\kern-.1667em\lower.7ex\hbox{E}\kern-.125emX}}
\begin{document}
\bstctlcite{IEEEexample:BSTcontrol}
\title{Impact of Hand Impairment and Occlusions on Hand Pose Estimation Accuracy in Augmented Reality Applications}
\author{Damian M. Manzone, Mathew Szymanowski, Olga Taran, Shuo Cai, Melissa Marquez-Chin, Tammy Zeng, Hardeep Singh, Cesar Marquez-Chin, and José Zariffa
\thanks{This work was supported by the Craig H. Neilsen Foundation (grant \#992531).}
\thanks{This work involved human subjects in its research. Approval of all
ethical and experimental procedures and protocols was granted by
the Research Ethics Board of the University Health Network under
Application No. 23-5248.}
\thanks{Damian M. Manzone is with the KITE Research Institute, Toronto Rehabilitation Institute, University Health Network, Toronto, ON M5G 2A2, Canada (e-mail: damian.manzone@uhn.ca). }
\thanks{Mathew Szymanowski and Melissa Marquez-Chin are with the Institute of Biomedical Engineering, University of Toronto (e-mail: m.szymanowski@utoronto.ca; melissa.marquezchin@mail.utoronto.ca).}
\thanks{Olga Taran was with KITE Research Institute, Toronto Rehabilitation Institute, University Health Network, Toronto, ON M5G 2A2, Canada. She is now with the Department of Health Sciences and Technology, ETH Zürich (e-mail: olga.taran@hest.ethz.ch).}
\thanks{Shuo Cai and Tammy Zeng were with the KITE Research Institute, Toronto Rehabilitation Institute, University Health Network, Toronto, ON M5G 2A2, Canada (e-mail: shuo.cai@mail.utoronto.ca; tammy.zeng3@uhn.ca).}
\thanks{Hardeep Singh is with the Department of Occupational Science \& Occupational Therapy and the Rehabilitation Sciences Institute, University of Toronto, Toronto, ON M5S 3G9, Canada (e-mail: hardeepk.singh@utoronto.ca).}
\thanks{Cesar Marquez-Chin is with the KITE Research Institute, Toronto Rehabilitation Institute, University Health Network, Toronto, ON M5G 2A2, Canada, and with the Institute of Biomedical Engineering, University of Toronto, Toronto, ON M5S 3G9, Canada (e-mail: cesar.marquezchin@utoronto.ca).}
\thanks{José Zariffa is with the KITE Research Institute, Toronto Rehabilitation Institute, University Health Network, Toronto, ON M5G 2A2, Canada, and also with the Institute of Biomedical Engineering, the Rehabilitation Sciences Institute, and the Edward S. Rogers Sr. Department of Electrical and Computer Engineering, University of Toronto, Toronto, ON M5S 3G9, Canada (e-mail: jose.zariffa@utoronto.ca).}}

\maketitle


\begin{abstract}
Mixed reality applications can be designed for hand rehabilitation. Augmented reality (AR) head mounted displays (HMDs) specifically allow for ecologically valid tasks because individuals can see their real environment and interact with real objects while receiving additional cues on the HMD. While these applications rely on accurate hand pose estimation, there is a gap in investigating the influence of hand impairment or occlusion from real-object interactions on pose estimation accuracy. Further, comparisons between AR HMD predictions and state-of-the-art pose estimation methods have not been established. The current study assessed pose estimation accuracy of the HoloLens~2 HMD and state-of-the-art pose estimation algorithms (WiLoR, HaMeR, WildHands, and MediaPipe) while individuals with cervical spinal cord injury (cSCI; n = 13, Neurological Level of Injury: C3–C6; American Spinal Injury Association Impairment Scale: A–D) and 15 uninjured controls interacted with clear and opaque objects. Ground truth estimates of 3D joint positions were generated via triangulation from a multi-camera setup. Pose estimation accuracy did not differ between the cSCI and uninjured control groups suggesting that 3D joint predictions from the HoloLens~2 and pose estimation algorithms can generalize to populations with hand impairment. Further, clear objects provided a small accuracy advantage over opaque objects ($\sim 0.1$~mm) and predictions from both WiLoR and HaMeR were slightly more accurate than the HoloLens~2 ($\sim 2$~mm). Overall, these results suggest that the HoloLens~2 may be viable for hand rehabilitation applications and the dataset generated can be used to refine pose estimation methods for hand-impaired populations.
\end{abstract}


\section{Introduction}
\label{sec:introduction}
\IEEEPARstart{C}{ervical} spinal cord injury (cSCI) results in impaired upper limb function, challenging daily interactions, and negative socio-economic impacts on the healthcare system and the individual~\cite{chan_lifetime_2019}. To improve hand function (a priority for this population~\cite{anderson_targeting_2004, samejima2026}), recent neurorehabilitation interventions have proposed the use of mixed reality technology and tools~\cite{de_araujo_efficacy_2019,scalise_virtual_2024,manzone2026augmented,gorman_use_2022,phan_effectiveness_2022}. Specifically, virtual and augmented reality (VR and AR) head mounted displays (HMDs) can collect large amounts of sensor data and present various types of information to users, but may rely on accurate predictions of hand joint positions from an egocentric perspective. However, in the specific context of rehabilitation, two factors can potentially limit the accuracy of these predictions. First, after a cSCI, individuals present distinct and variable hand postures and grasping patterns compared to uninjured individuals~\cite{dolbow_electrical_2023,dousty_grasp_2023,dousty2024personalized}. Second, hand function rehabilitation relies on individuals reaching, grasping, and lifting real objects~\cite{waddell2017,behrman_activity-based_2017}. Grasping real objects can pose a specific challenge for pose estimation methods because the object can block or occlude vision of parts of the fingers from the camera's egocentric perspective~\cite{li2026challenges}. Thus, assessing pose estimation accuracy of mixed reality systems during real-object interactions is a necessary step in the development of rehabilitation tools for individuals with cSCI.

The impact of hand impairment on pose estimation has not yet been well established. To quantify pose estimation accuracy of mixed reality HMDs, estimates of hand joint positions are compared to ground truth estimates. For uninjured participants, errors between onboard hand tracking from an HMD and ground truth estimates range from approximately 3--20~mm (e.g., Meta's Quest~2~\cite{abdlkarim_methodological_2024}, HTC's Vive Lighthouse Tracking System~\cite{ikbal_dynamic_2021}, Microsoft's HoloLens~2~\cite{schneider_accuracy_2021,soares_accuracy_2021,bertolasi_evaluation_2025}). However, from the few studies that employed users with hand impairment (e.g.,~\cite{bazyk_validation_2025,casile_quantitative_2023}), results are mixed when quantifying differences of HMD hand tracking accuracy between individuals with and without hand impairment. For example, the HoloLens~2 was able to accurately predict tap frequency and amplitude when compared to a motion capture system, but only for uninjured participants and not individuals with Parkinson's disease~\cite{bazyk_validation_2025}. On the other hand, Meta's Quest~2 was able to accurately predict hand position and velocity in both paretic and non-paretic limbs in individuals living with stroke when compared to estimates from a marker-based system~\cite{casile_quantitative_2023}. Both of the aforementioned studies, however, did not involve any object interactions. Thus, there is a gap in understanding pose estimation accuracies of HMDs during object interactions for users with hand impairment.

Object interactions are of central importance in the context of rehabilitation, where applications that prioritize real-world functionality and task specificity can contribute to improved recovery~\cite{waddell2017}. For example, employing AR systems specifically may be advantageous for rehabilitation because individuals can see their real hand in the real environment while interacting with real objects. Compared to virtual objects, real objects provide additional touch, weight and proprioceptive information that match interactions during functional activities of daily living~\cite{manzone2026augmented,behrman_activity-based_2017,kleim_principles_2008}. When interacting with real objects, however, the finger and joint locations from the HMD's egocentric perspective can be occluded by the object and cause inaccurate joint position estimates~\cite{myanganbayar_partially_2019}. To quantify pose estimation accuracy during object interactions, a recent study had uninjured participants hold different objects statically while wearing a HoloLens~2~\cite{bertolasi_evaluation_2025}. Compared to the marker-based ground truth estimates, mean angular errors within each joint ranged from under $1^\circ$ to approximately $22^\circ$~\cite{bertolasi_evaluation_2025}. However, having uninjured individuals interact with objects in a static manner may not generalize well to rehabilitation contexts in which individuals with hand impairment would interact with objects dynamically (i.e., reaching, grasping, and lifting). Further, although Bertolasi and colleagues~\cite{bertolasi_evaluation_2025} used opaque and clear or transparent objects, object transparency was not systematically manipulated. That is, opaque and clear objects differed in number, size, and shape, limiting interpretations of the influence of occlusion on pose estimation accuracy.

One final important consideration is the pose estimation accuracy of HMDs relative to other state-of-the-art methods. That is, if accuracies of the state-of-the-art methods outperform onboard tracking systems, they can potentially be integrated with the HMD technology to improve the user experience and clinical usefulness of the application. Recent machine learning methods have demonstrated high 3D pose estimation accuracy on public datasets~\cite{potamias2025wilor,pavlakos2024reconstructing}, but to our knowledge, no previous study has directly compared their accuracies to predictions from an AR HMD. To do so, datasets would require recordings directly from the egocentric perspective of the AR HMD along with joint position predictions from the onboard hand tracking system. However, these specific types of datasets (i.e., HMD recordings with onboard predictions) are not represented in existing benchmarks of publicly available egocentric hand pose datasets~\cite{li2026challenges, fan2024benchmarks}. In the context of rehabilitation, it is advantageous for joint position predictions to be generated as quickly as possible (i.e., real-time), so it is also important to compare pose estimation methods that are designed to be lightweight and run in real-time (e.g.,~\cite{potamias2025wilor}).

The objective of the present study was to examine the impact of and interactions between hand impairment, occluded views and pose estimation algorithms on hand pose estimation accuracy in egocentric views from an AR HMD. Preliminary findings from these experiments found no differences between individuals with and without cSCI but reduced accuracy when comparing large versus small objects~\cite{Manzone_embc_2026}, however focused only on the HoloLens~2 hand tracking. By including additional pose estimation approaches and comparing opaque versus clear objects, we can fully elucidate the interactions between factors leading to the following contributions: 1) further characterizing the impact of tetraplegia on pose estimation accuracy during real-object interactions, 2) characterizing the impact of occlusion on hand pose estimation, 3) characterizing differences in pose estimation accuracy across different pose estimation methods, and 4) identifying interactions between the aforementioned factors (e.g., does one pose estimation method better account for occlusions or hand impairments than others?). Specifically, individuals with cSCI and uninjured controls wore the HoloLens~2 HMD while reaching, grasping and lifting different objects. To further understand the impact of occlusion on pose estimation accuracy, participants interacted with either opaque or clear versions of the same object to minimize grasping variability between occlusion conditions. Lastly, joint position estimates directly from the HoloLens~2 along with other state-of-the-art pose estimation algorithms were compared to ground truth estimates generated from a multi-camera setup. Based on our previous results, we hypothesized no accuracy differences between uninjured and cSCI groups, at least for the HoloLens~2 predictions. Further, we hypothesized that pose estimation accuracy would be higher when interacting with clear objects and that state-of-the-art pose estimation methods would yield more accurate predictions than onboard predictions from the HoloLens~2.

\section{Methods}
\label{sec:methods}

\subsection{Participants}

Thirteen individuals with cSCI (10 Male, 3 Female; Age: $49 \pm 13$ years; Neurological Level of Injury: C3--C6; AIS: A--D; see~\autoref{table:participants} for full demographic details) and 15 uninjured controls (8 Male, 7 Female; Age: $30 \pm 6$ years) were included in the analyses for the study. Participants self-reported sex and gender separately, with all male participants reporting their gender as man and all female participants reporting their gender as woman. Two additional individuals participated in the study (1 Male -- Age: 38, Neurological Level of Injury: C5, AIS: C; 1 Female -- Age: 33, Neurological Level of Injury: C5, AIS: A) but were excluded from the analyses because they could not keep their hands or the camera calibration cube (see \autoref{sec:exp_setup}) in the field of view of the camera for the majority of the frames during object interactions. This meant that estimates of joint positions could not be generated for the majority of the frames. All participants provided written informed consent and all protocols were approved by the Research Ethics Board of the University Health Network (\#23-5248).

\begin{table}[!tb]
\centering
\caption{Participant Demographics for the cSCI group.\\
\textit{Note: T = Traumatic, NT = Non-traumatic.}} 
\label{table:participants}
\setlength{\tabcolsep}{4pt} 
\begin{tabular}{lcccccc}
\toprule
Participant & \makecell[c]{Age \\ (years)} & Sex & \makecell[c]{Injury \\ level} & \makecell[c]{AIS \\ grade} & \makecell[c]{Type of \\ injury} & \makecell[c]{Months since \\ injury} \\
\midrule
1 & 65 & M & C6 & D & T & 5 \\
3 & 64 & F & C3 & D & NT & 21 \\
4 & 31 & M & C3 & D & NT & 2 \\
5 & 52 & M & C4 & D & T & 160 \\
6 & 28 & M & C6 & B & T & 68 \\
7 & 31 & M & C6 & D & NT & 2 \\
8 & 59 & F & C4 & D & NT & 74 \\
9 & 49 & M & C4 & B & NT & 51 \\
10 & 39 & M & C4 & C & T & 47 \\
11 & 54 & M & C4 & C & NT & 34 \\
12 & 56 & M & C3 & D & NT & 56 \\
13 & 61 & M & C4 & A & NT & 28 \\
15 & 52 & F & C6 & B & T & 36 \\ 
\midrule
Mean$\pm$SD & 49$\pm$13 & & & & & 45$\pm$42 \\ 
\bottomrule
\end{tabular}
\end{table}

\subsection{Experimental Setup}
\label{sec:exp_setup}
\begin{figure}[!t]
\centerline{\includegraphics[width=\columnwidth]{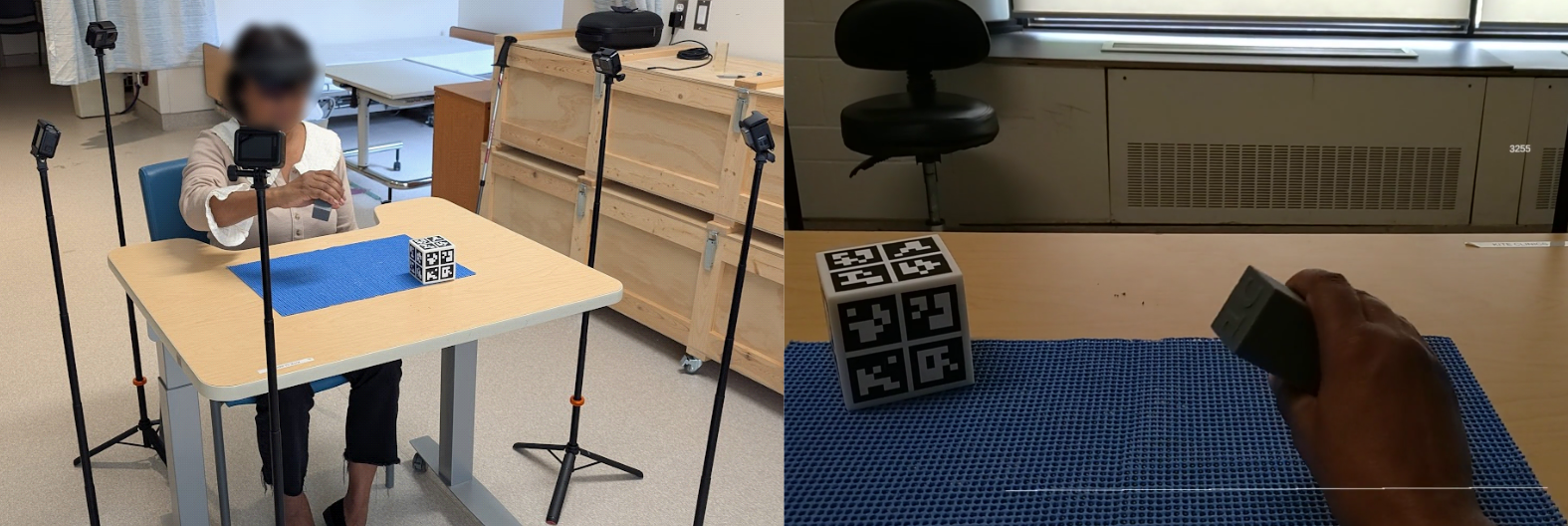}}
\caption{\textbf{Left Panel}: Depiction of the experimental setup with the participant lifting a block while wearing the HoloLens~2 AR headset and surrounded by five cameras. \textbf{Right Panel}: Depiction of the participant's view from the headset's egocentric perspective. The white line displayed the bottom edge of the recording window or field of view of the HMD.}
\label{fig:setup}
\end{figure}

All participants sat at a table with either their mobility device or chair while wearing a Microsoft HoloLens~2 HMD (Redmond, WA, USA; see~\cite{ungureanu2020hololens} for a description of the integrated hardware and software). Five cameras (GoPro Hero5; San Mateo, CA, USA) attached to tripods surrounded the participant and table and were used to record grasping interactions from multiple perspectives (see~\autoref{fig:setup}). The camera behind the participant was placed behind the shoulder/arm performing the movements for a given trial, but the other four cameras were fixed. The five tripod cameras and the AR HMD recording interactions from an egocentric perspective all recorded at 30 frames per second. A 3D printed cube with ArUco markers on all sides~\cite{garrido-jurado_automatic_2014} was also placed at the top corner of the table on the opposite side of the moving limb. The cube was used to calculate intrinsic and extrinsic camera parameters, and all cameras were adjusted to ensure sight of the cube at all times.

\begin{figure}[!b]
\centering
\includegraphics[width=0.9\columnwidth]{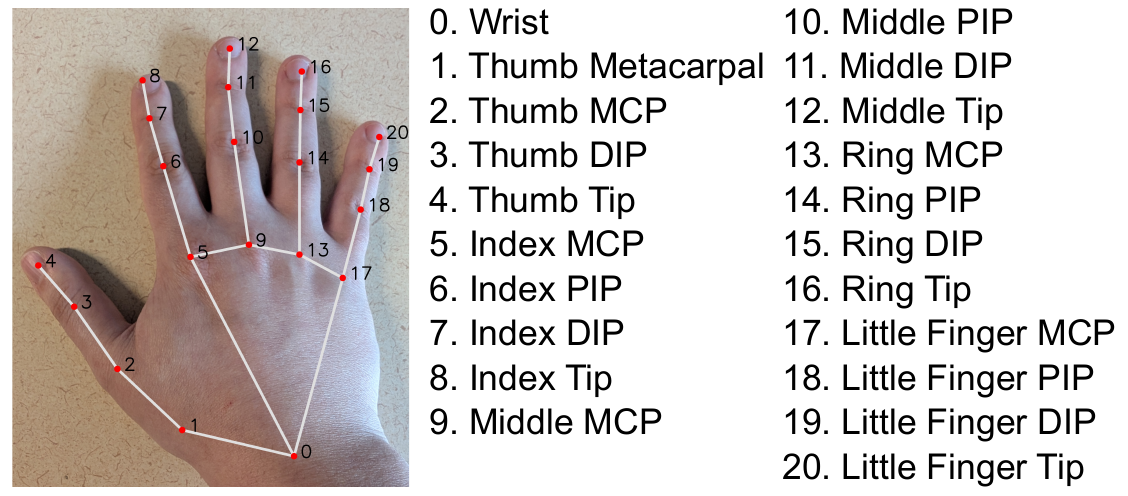}
\caption{Depiction of the 21 joints estimated by the HoloLens~2 and included in the ground truth estimation. MCP = metacarpophalangeal, PIP = proximal interphalangeal, DIP = distal interphalangeal.}
\label{fig:joints}
\end{figure}

A custom-made Unity (Unity Technologies, San Francisco, CA, USA) application was created for the HoloLens~2. The application recorded 3D estimates of 21 joints for either the left or right hand for every frame using the HMD's onboard hand tracking system via the Mixed Reality Toolkit~2 (see~\autoref{fig:joints}). A small white number indicating the frame number was also used to synchronize joint estimates across cameras frames. Lastly, a white line showed the edge of the HMD's field of view and participants were instructed to try and keep their hands above the line during object interactions (see~\autoref{fig:setup}). 

\subsection{Procedure}
\begin{figure}[!b]
\centering
\includegraphics[width=0.9\columnwidth]{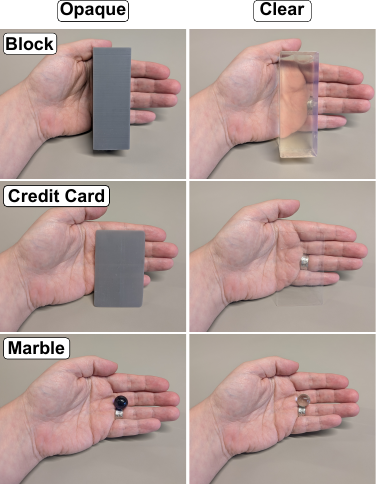}
\caption{Depiction of the opaque and clear versions of the block, credit card, and marble that participants interacted with.}
\label{fig:objects}
\end{figure}

Prior to beginning experimental trials and wearing the HoloLens~2, participants with cSCI completed sensory and motor assessments using the Graded Redefined Assessment of Strength Sensibility and Prehension (GRASSP version 1~\cite{kalsi-ryan_graded_2012}). After completing the assessment the experiment began by wearing the HoloLens~2, starting the application and recordings from all six cameras. A tablet was placed on the table in front of participants and played a video displaying a different colour screen every second, which was used to synchronize the cameras offline (see \autoref{sec:ground_truth_methods}). A recording session was conducted for object interactions with the right and left hand separately. After the experimenter placed an object in front of the participant, they were required to reach, grasp and lift the object about 10~cm off the table followed by pronating and supinating their wrist and placing the object back onto the table. Participants interacted with six objects, three times each, for a total of 18 trials per hand. The six objects included an opaque and clear version of a block, credit card, and a marble (see~\autoref{fig:objects}). The three objects were specifically chosen because they require different types of grasp used in functional assessments for people with cSCI (i.e., marbles require tip pinch grasps, blocks require cylindrical grasps, and credit cards require lateral key pinch grasps). It is worth noting, however, that participants were instructed to interact with objects naturally without providing explicit requirements for grasping posture. The blocks and credit cards were 3D printed from the Toronto Rehabilitation Hand Function Test~\cite{kapadia_3-dimensional_2021,kapadia_preliminary_2021}. The opaque versions used FormLabs Gray Resin V4 and the clear versions used FormLabs Clear Resin V4 and applied a clear spray coating after sanding the object to achieve the clear finish. Thus, all objects were approximately the same size and weight, with the only differentiating feature being their transparency.  

\subsection{Ground Truth Estimation}
\label{sec:ground_truth_methods}
The following steps were taken to generate the 3D ground truth estimates for the 21 joint positions:
\begin{enumerate}
    \item \textbf{Splitting frames:} Videos from the five tripod cameras were split into individual image frames.
    \item \textbf{Synchronization:} The first frame in which a green colour screen on the tablet was detected in each camera was used to synchronize frames across cameras.
    \item \textbf{Frame Selection:} The start of each trial was defined as approximately 10 frames prior to object contact (i.e., while the participant approached the object) and the end of the trial was defined as 10 frames after the participant released the object after placing it back onto the table. This was detected manually and from the egocentric camera's perspective. If the hand was not within the camera's field of view within the 10 frames, the first frame in which the hand entered the field of view prior to object contact or last frame before the hand exited the field of view were used as start and end, respectively.
    \item \textbf{Estimating camera parameters:} Intrinsic and extrinsic camera parameters of each camera were calculated using the estimated coordinates of the ArUco cube and its measured 3D real-world coordinates.
    \item \textbf{Estimating 2D joint positions:} Estimates of 2D joint positions for each of the 21 joints articulations and for each camera and for each frame were generated. Bounding boxes were generated using Google's MediaPipe Hands~\cite{zhang_mediapipe_2020} and joint position estimates were estimated using HRNetV2~\cite{wang_deep_2020}. HRNetV2 was chosen because it outperformed other pose estimation methods when benchmarking 2D egocentric datasets (see \cite{taran2025benchmarking}).
    \item \textbf{Estimating 3D joint positions:} 3D joint coordinates were estimated by triangulating the 2D joint coordinates from each camera. Theoretically, this would produce 21, 3D joint position estimates for each frame but four criteria were used to determine if the 2D estimates from a specific camera could be included in the triangulation or if the entire frame needed to be excluded from generating 3D joint estimates:
    \begin{enumerate}
        \item \textit{Bounding box error:} If a bounding box was not detected a 2D estimate could not be generated (e.g., if the hand was outside of the camera's field of view).
        \item \textit{Camera parameter error:} If the camera parameters could not be estimated or if the estimated yielded a high reprojection error (i.e., root mean square distance between the detected 2D cube marker coordinates and the reprojected coordinates exceeding 3.5 pixels), the specific camera's frame was excluded. These errors may occur if the cube was occluded, out of the camera's field of view, or if the cube was blurred in the image.
        \item \textit{Random sample consensus:} Estimates from each camera were considered outliers and excluded from triangulation based on random sample consensus (RANSAC; parameters: confidence threshold = 0.1, reprojection epsilon = 100, number of iterations = 25).
        \item \textit{Number of cameras:} If less than two cameras were included in the triangulation, 3D joint position estimates could not be generated for that frame.
    \end{enumerate}
    \item \textbf{Ground truth outlier analyses:} The following outlier analyses were conducted to ensure the quality of the ground truth estimates:
    \begin{enumerate}
        \item \textit{Absolute differences between frames:} If the Euclidean distance between each joint’s position in consecutive frames exceed 10~cm in one frame, the frame was discarded.
        \item \textit{Relative differences between frames:} the average and standard deviation of the distance traveled within each joint and finger segment lengths were calculated over a 10-frame, non-overlapping window. If any distance or finger segment length exceeded 2.5 standard deviations of the mean, the frame was discarded. This continued for every non-overlapping 10-frame window.
    \end{enumerate}
In total, 60,128 out of a possible 159,167 frames were included (i.e., approximately 38\%) which represents a conservative estimate of the ground truth joint positions. All 3D ground truth estimates, 3D estimates from the HoloLens~2, and individuals image frames that were included in the analysis are publicly available (see \autoref{sec:data_availability}).

\end{enumerate}

\subsection{Pose Estimation Algorithm Selection}

Videos from the egocentric perspective of the HoloLens~2 were also split into individual video frames and synchronized with the five tripod cameras. For each frame, 3D joint position estimates were generated from the HoloLens~2 (i.e., via its onboard articulated hand tracking system~\cite{ungureanu2020hololens}) and from four other pose estimation algorithms (i.e., using the individual frame image). The four algorithms were WiLoR~\cite{potamias2025wilor}, HaMeR~\cite{pavlakos2024reconstructing}, MediaPipe Hands~\cite{zhang2020mediapipe}, and WildHands~\cite{prakash20243d}. Algorithms were selected based on three criteria: (1) reported state-of-the-art or competitive performance on public 3D hand pose datasets, (2) publicly available source code and pre-trained model weights with a functional demo, and (3) relevance to egocentric hand tracking during object interaction.
WiLoR~\cite{potamias2025wilor} and HaMeR~\cite{pavlakos2024reconstructing} were included as recent transformer-based methods reporting state-of-the-art accuracy on public benchmarks. WiLoR was additionally designed for real-time, lightweight deployment. WildHands~\cite{prakash20243d} was included for its focus on egocentric, in-the-wild hand-object interactions, and MediaPipe Hands~\cite{zhang2020mediapipe} as a widely adopted, computationally efficient baseline. It is worth noting that for some frames the algorithms could not generate a 3D prediction of the joint positions. While WiLoR, HaMeR, and WildHands could not generate a predictions on $<$1\% of the accepted ground truth frames, MediaPipe could not generate a prediction on $\sim 22$\% of the accepted ground truth frames.

\begin{figure*}[htbp]
\centering
\includegraphics[width=0.95\textwidth]{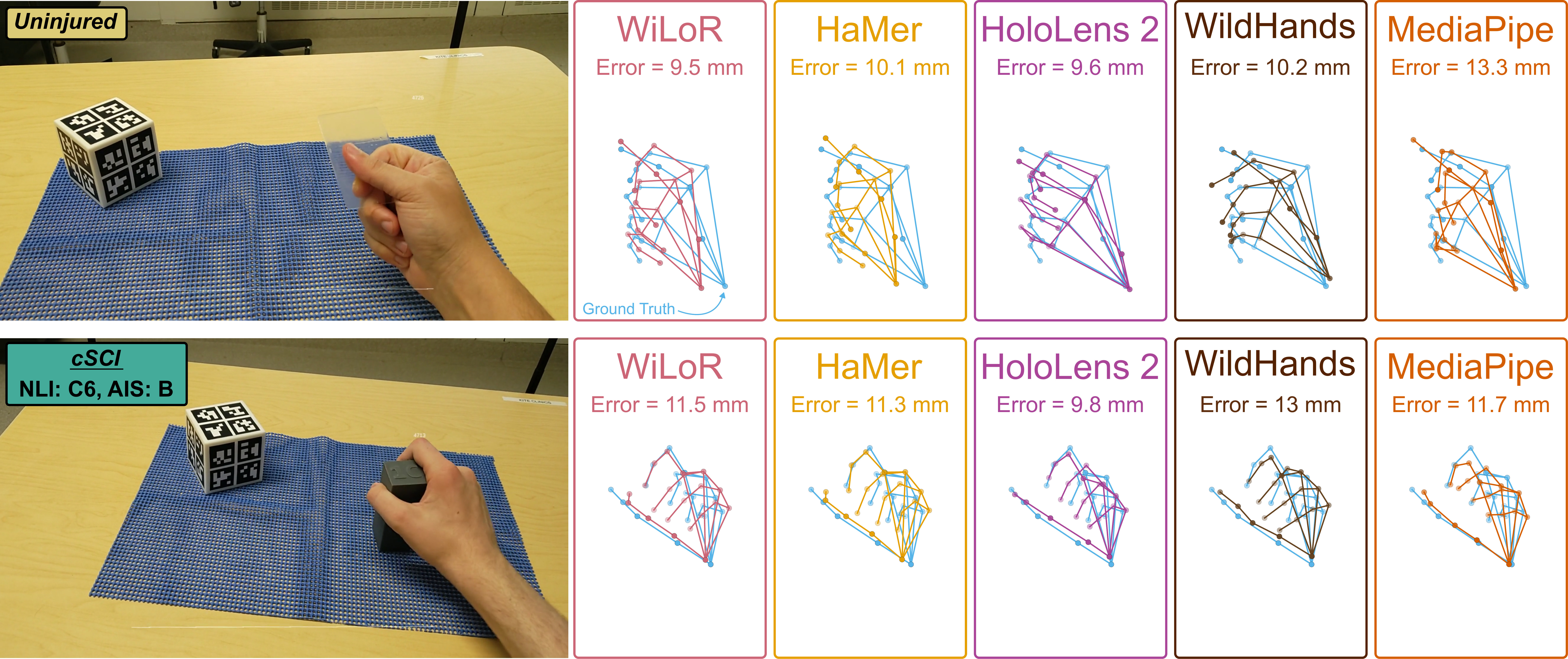}
\caption{Example of comparisons between the 3D joint positions estimated for the ground truth and each algorithm for one frame when a uninjured participant was grasping the clear credit card (top row) and when a cSCI participant was grasping the opaque block (bottom row). The Procrustes-Aligned Mean Per Joint Error for that specific frame and algorithm is displayed below each comparison. Ground truth from the multi-camera triangulation is shown in blue.}
\label{fig:algorithm_comparison}
\end{figure*}

\subsection{Data Analyses}
\subsubsection{Main Analyses}
Procrustes-aligned~\cite{gower1975} mean per-joint position error~\cite{kanazawa2018} (PA-MPJPE) was used to compare 3D joint pose estimates generated by all algorithms and the ground truth estimates. After rigid alignment, PA-MPJPE measured the difference between an estimated position and a ground truth position averaged across all joints for each frame (see~\autoref{fig:algorithm_comparison}). Then, the PA-MPJPE was averaged across all frames for a given trial, followed by averaging across all trials for a given condition for each algorithm. This process resulted in one final PA-MPJPE value for each condition (e.g., the HoloLens~2, clear object, cSCI condition). PA-MPJPEs were submitted to a 5 algorithm (HoloLens~2, WiLoR, HaMeR, MediaPipe, WildHands) by 2 transparency (clear, opaque) by 2 group (uninjured, cSCI) mixed measures ANOVA with group the between subjects factor and algorithm and transparency within-subjects factors. Alpha was set at $p = 0.05$ for all main effects and interactions. Significant main effects and interactions were decomposed using Bonferroni corrected paired samples t-tests, adjusting the alpha value according to the number of comparisons.

To understand if the errors measured by each algorithm were related to the level of hand impairment in the cSCI group, the left and right hand Prehension Performance Subscores from the GRASSP were correlated with the total average PA-MPJPE per hand for each algorithm using separate Spearman's rank order correlations.

\subsubsection{Exploratory Analyses}
Exploratory analyses were conducted to understand if specific objects influenced occlusion effects and if the participant's sex influenced pose estimation accuracy overall. For the object analysis, three Bonferroni corrected preplanned paired sample t-tests compared clear-opaque differences within each object. For the sex analysis, a preplanned independent samples t-test was conducted for the overall PA-MPJPEs between males and females (i.e., collapsed across algorithm and transparency).

\section{Results}
\label{sec:results}
\subsubsection{Main Analyses}
The mixed measures ANOVA yielded significant main effects of algorithm ($F(4, 104) = 169.82$, $p < 0.001$, $\eta_{p}^{2} = 0.87$) and transparency ($F(1, 26) = 4.65$, $p = 0.04$, $\eta_{p}^{2} = 0.15$), but did not yield a significant main effect of group ($F(1, 26) = 0.80$, $p = 0.38$, $\eta_{p}^{2} = 0.03$). Further, the only significant interactions were a two-way interaction between algorithm and transparency ($F(4, 104) = 8.71$, $p < 0.001$, $\eta_{p}^{2} = 0.25$) and a three-way interaction between algorithm, transparency, and group ($F(4, 104) = 3.85$, $p < 0.01$, $\eta_{p}^{2} = 0.13$).

\begin{figure}[!b]
\centering
\includegraphics[width=0.9\columnwidth]{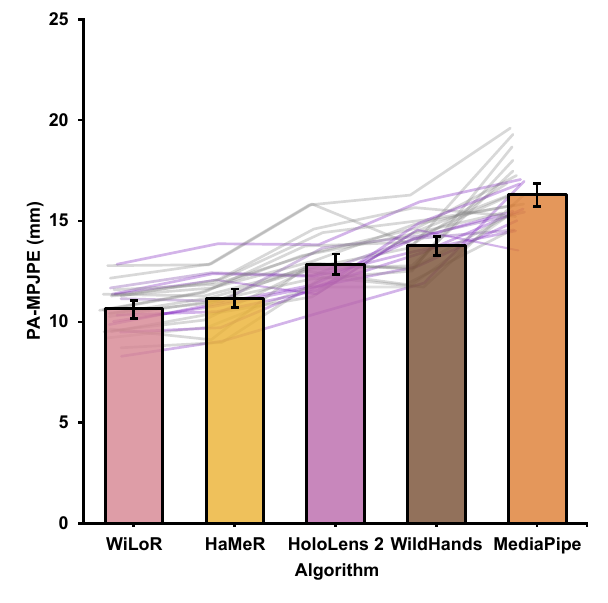}
\caption{The significant main effect of algorithm. All comparisons between algorithms were significantly different (i.e., all $p < 0.005$). Individual participants are shown as lines behind the bars. Grey lines are male participants and purple lines are female participants. Error bars represent 95\% confidence intervals.}
\label{fig:me_algorithm}
\end{figure}

\begin{figure}[htbp]
\centering
\includegraphics[width=0.9\columnwidth]{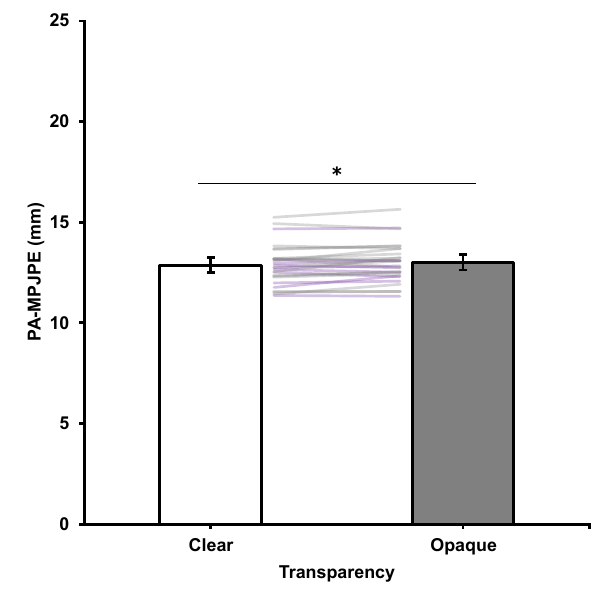}
\caption{The significant main effect of transparency. Individual participants are shown as line between the bars, averaged across all groups and algorithms. Grey lines are male participants and purple lines are female participants. Error bars represent 95\% confidence intervals.}
\label{fig:me_transparency}
\end{figure}

\begin{figure*}[htbp]
\centering
\includegraphics[width=0.9\textwidth]{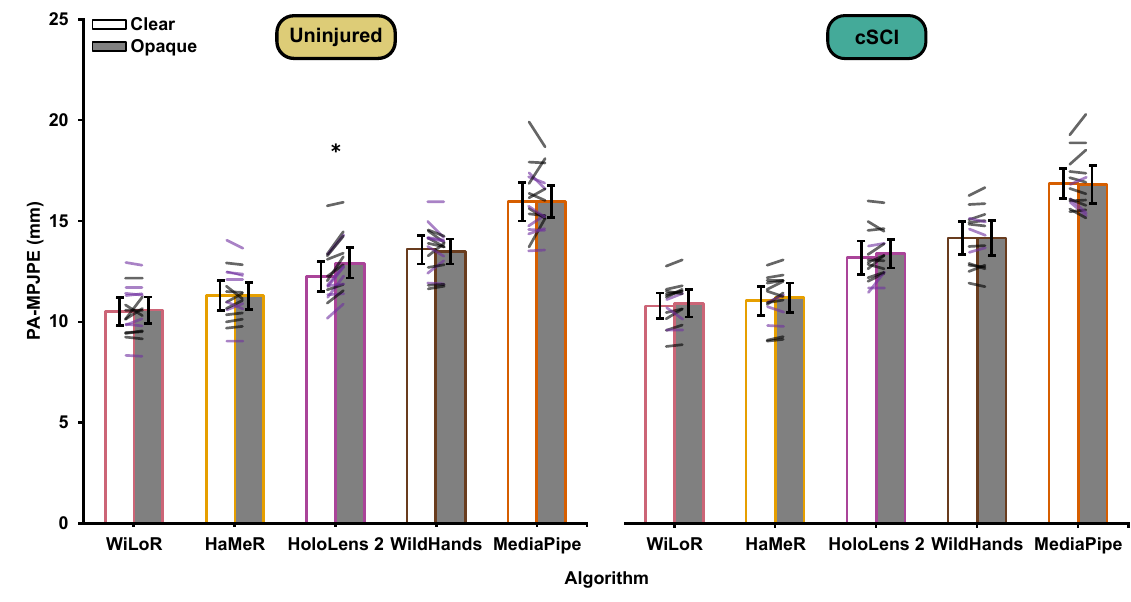}
\caption{The significant group by transparency by algorithm interaction. After correction for multiple comparisons, the only clear-opaque difference was in the uninjured group with the HoloLens~2 algorithm ($p < 0.005$). Individual participants are shown as lines between bars within each algorithm. Grey lines are male participants and purple lines are female participants. Error bars represent 95\% confidence intervals.}
\label{fig:3way_interaction}
\end{figure*}

The insignificant main effect of group suggests that there was no difference in pose estimation accuracy between the uninjured and cSCI groups (see~\autoref{fig:me_group}). For the main effect of algorithm, ten paired samples t-tests were conducted to compare each algorithm to each other (Bonferroni corrected $\alpha = 0.005$). All pairwise comparisons were significant, with the WiLoR having the lowest error followed by HaMeR, HoloLens~2, MediaPipe and WildHands (see~\autoref{fig:me_algorithm}). For the main effect of transparency, grasping opaque objects ($13.0 \pm 1$~mm) led to higher error overall than grasping clear objects ($12.9 \pm 1$~mm; see~\autoref{fig:me_transparency}). For the interaction between algorithm and transparency, the magnitude of the difference in error between the opaque and clear objects was compared across the five algorithms (i.e., ten paired samples t-tests total, Bonferroni corrected $\alpha = 0.005$). From the comparisons, the magnitude of the opaque-clear difference with the HoloLens~2 was larger than with all other algorithms (i.e., higher error with the opaque condition; all $p < 0.005$). For the three-way interaction between algorithm, transparency and group, ten paired samples t-tests were used to compare opaque and clear conditions within each algorithm and for both groups (e.g., opaque vs. clear for HoloLens~2 in uninjured group, opaque vs. clear for HoloLens~2 in cSCI group, etc.; see~\autoref{fig:3way_interaction}). From the comparisons, the only significant difference was a larger error when grasping the opaque object than the clear object when error was measured with the HoloLens~2 for the uninjured group only (opaque: $12.9 \pm 1.4$~mm; clear: $12.3 \pm 1.4$~mm; $p < 0.005$).

\begin{figure*}[!b]
\centering
\includegraphics[width=0.9\textwidth]{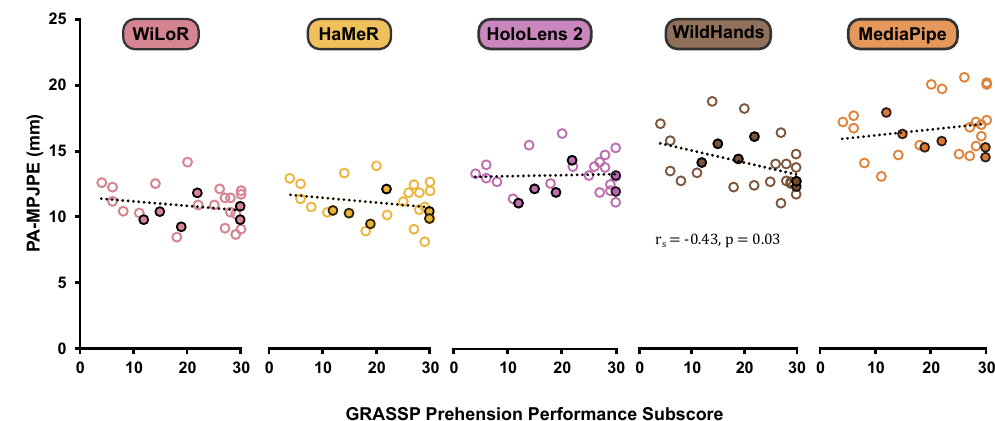}
\caption{The correlations between error and Prehension Performance Subscores separated by algorithm. Each participant has a data point for average error and GRASSP Prehension Performance Subscore for their left and right hand separately. Filled circles represent female participants. The WildHands algorithm was the only algorithm that yielded a significant correlation.}
\label{fig:correlation}
\end{figure*}

The correlational analyses between PA-MPJPE and GRASSP Prehension Performance Subscore yielded a significant moderate negative correlation for the WildHands algorithm only ($r_s(24) = -0.43$, $p = 0.03$; all other $p > 0.25$). The negative correlation suggests that pose estimation accuracy decreased with higher levels of hand impairment when joint predictions were made with the WildHands algorithm (see~\autoref{fig:correlation}).

\subsubsection{Exploratory Analyses}
\begin{figure}[!t]
\centering
\includegraphics[width=0.9\columnwidth]{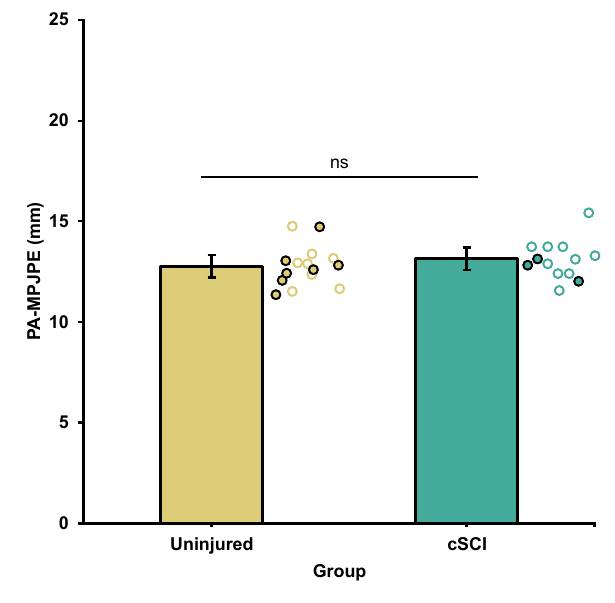}
\caption{The insignificant main effect of group. Individual participants are shown as circles beside the bar which represents the group mean, averaged across all objects and algorithms. Filled circles represent female participants. Error bars represent 95\% confidence intervals. ns = not significant.}
\label{fig:me_group}
\end{figure}

For the exploratory object analysis, the clear-opaque difference was only significant for the credit card, but not the block or marble ($p < 0.01$; clear credit card: $13.1 \pm 1.1$~mm; opaque credit card: $13.5 \pm 1.2$~mm). For the exploratory sex analysis, the independent samples t-test did not yield a significant difference between male and female participants for PA-MPJPE overall ($p = 0.38$; see~\autoref{fig:me_group} for a sex breakdown within the main analyses).

\section{Discussion}
\label{sec:discussion}

The goal of the current study was to assess pose estimation accuracy of both onboard HMD and offline pose estimation methods when individuals with cSCI interacted with opaque and clear objects. In line with our hypotheses, there were no differences in pose estimation accuracy when individuals with cSCI or uninjured controls reached, grasped and lifted different objects. Moreover, errors were consistently similar in both groups regardless of the pose estimation methods used to predict joint positions. Further, joint position predictions were significantly more accurate when interacting with clear compared to opaque objects, but the overall difference was quite small (i.e., 0.1~mm). Lastly, state-of-the-art pose estimation methods (i.e., WiLoR and HaMeR) outperformed predictions from the HoloLens~2 HMD. 

Compared to previous work, this is the first study to compare HMD pose estimation accuracy when individuals with hand impairment interacted with real objects. The pose estimation accuracies reported in the current study (i.e., overall mean = 13~mm) are also within the range reported in the literature that employs the HoloLens~2 HMD (i.e., errors in the literature range from 3--20~mm; see~\cite{schneider_accuracy_2021,soares_accuracy_2021,bertolasi_evaluation_2025}). This work extends previous HMD work that had participants perform target reaching~\cite{abdlkarim_methodological_2024} and tracing tasks~\cite{schneider_accuracy_2021,bertolasi_evaluation_2025} to include dynamic interactions with objects. That is, the previous study which included object interactions had uninjured participants hold objects in a still position and only measured differences in joint angles from HoloLens~2 and the ground truth estimates. In the current study, however, pose estimation accuracy was assessed throughout all stages of an object interaction (i.e., reaching, grasping, lifting, rotating, and placing the object back down) which better reflects real use cases within and beyond rehabilitation.

Within the context of rehabilitation, it is also important to consider the individuals using rehabilitation applications and how the tasks they perform may influence pose estimation accuracy. For example, when creating applications that incorporate pose estimation methods, it is first important to assess pose estimation accuracy in the intended population (e.g., individuals with cSCI;~\cite{stenum_applications_2021}). In line with preliminary analyses~\cite{Manzone_embc_2026}, there were no significant differences in pose estimation accuracy overall when individuals with cSCI or the uninjured controls interacted with the objects and this pattern was similar across all other algorithms tested. Further, the exploratory sex analysis revealed no pose estimation accuracy differences between male and female participants. Although differences between groups was not significant overall, it is worth noting that for one of the five algorithms tested (i.e., WildHands), the level of pose estimation accuracy was significantly related to the level of impairment. Thus, this relationship provides some evidence that hand impairment can influence pose estimation accuracy with lower accuracy levels related to higher impairment levels. It is worth noting, however, that the difference in errors between the participant with the highest level of hand impairment (i.e., 4 on GRASSP Prehension Performance Subscore) and the participants with the lowest level of hand impairment (i.e., 30 on the GRASSP Prehension Performance Subscore) was only $\sim 4$~mm. Thus, when considering the small error differences across levels of hand impairment and the lack of significant relationships with the three other algorithms and the HoloLens~2 HMD, the findings suggest that pose estimation predictions may be generalizable to the potentially unique hand postures employed by individuals with cSCI (e.g.,~\cite{dolbow_electrical_2023,dousty2024personalized}).

Interacting with real objects, which is ecologically valid for rehabilitation, also introduces potential issues of finger occlusion which can limit pose estimation accuracy~\cite{myanganbayar_partially_2019}. Using clear and opaque versions of the same object improved on previous work because it isolated the influence of occlusion on pose estimation accuracy. That is, previous work included different objects that were either clear or opaque~\cite{bertolasi_evaluation_2025}, so the work could not isolate the influence of object from the influence of occlusion. In the current study, although there was an overall accuracy advantage when interacting with clear objects, exploratory analyses revealed that the effect was driven by differences when comparing clear and opaque credit cards. Potential reasons for lack of differences with the marble and block may relate to the small size of the marble and the additional refractions caused by the shape of the block which may have changed the visual position of the finger (see~\autoref{fig:objects}). However it is important to consider the magnitude of the improvement when interacting with clear objects (i.e., $\sim 0.1$~mm accuracy improvement). Although the objects were chosen because of their relevance to therapeutic tasks, the small size of the objects relative to previous work (i.e.,~\cite{myanganbayar_partially_2019}) may account for the smaller differences in the current study. Thus, future work can systematically employ larger amounts of different sized and shaped objects to understand potential advantages of using clear objects for pose estimation accuracy. 

The methodology employed in the current study (i.e., recording egocentric video from the HoloLens~2 HMD) also allowed pose estimation comparisons between onboard HMD hand tracking and offline state-of-the-art pose estimation methods. This comparison was relevant to assess if other pose estimation methods outperform the onboard hand tracking of the HoloLens~2 on therapy-relevant tasks. Overall, estimates from WiLoR~\cite{potamias2025wilor} and HaMeR~\cite{pavlakos2024reconstructing} outperformed those generated from HoloLens~2 and all other algorithms tested. Further, the pattern of results from offline pose estimation methods was similar to the HoloLens~2 (cf.~\autoref{fig:3way_interaction}). This may reflect differences in model capacity and training data scale: WiLoR and HaMeR both employ large vision transformer backbones trained on millions of in-the-wild hands, with WiLoR additionally incorporating a coarse-to-fine refinement module that resamples image aligned features to correct pose misalignment~\cite{potamias2025wilor}. By contrast, the HoloLens~2's onboard tracker and MediaPipe are both optimized to run on-device in real-time~\cite{ungureanu2020hololens,zhang2020mediapipe}, which constrains model size and training data, while WildHands uses a smaller ResNet-50 backbone trained on a narrower set of egocentric datasets~\cite{pavlakos2024reconstructing}. It is also worth considering that the HoloLens~2 utilized a depth camera for joint position predictions whereas the other algorithms relied on RGB egocentric images and may account for slight differences in pose estimation accuracy (i.e., clear-opaque difference with the HoloLens~2 only; see~\autoref{fig:3way_interaction}). With respect to the dataset generated from the current study, it is the first ground truth annotated egocentric dataset which includes object interactions and both a group of individuals with hand impairment via cSCI and a group of uninjured controls (see~\cite{taran2025benchmarking,li2026challenges,banerjee2025hot3d} for discussions of other public egocentric datasets). That is, while other egocentric datasets exist showcasing object interactions from individuals with hand impairment (e.g., \cite{nguyen2025rehabhand}), this is the first to include annotated ground truth 3D hand pose estimation. Thus, future work can use the dataset to assess new or existing pose estimation methods for applications in clinical populations~\cite{stenum_applications_2021} and compare them to predictions from an existing AR HMD. 

When interpreting the results of the current study, the magnitude of pose estimation accuracy errors requires careful consideration. First, future work is necessary to understand whether the 13~mm error from the HoloLens~2 is acceptable for clinical applications. For example, when shown a real-time depiction of the onboard hand tracking while assessing a rehabilitation application, 80\% of participants with cSCI agreed or strongly agreed that the hand tracking seemed accurate~\cite{manzone2026augmented}. However, the acceptability of error likely relates to the degree of precision required for the specific application or clinical assessment. Second, although overall improvements were shown with the state-of-the-art pose estimation methods, an open question concerns whether a $\sim 2$~mm improvement in overall accuracy is worth potential practical limitations (see~\autoref{fig:algorithm_comparison} for a visual depiction of errors). For example, the HMD's processing power may limit the integration of other pose estimation methods which could limit the ability to perform real-time joint position predictions unless another computer is in the loop. Thus, future work should gather perspectives from end users and clinicians in addition to assessing how pose estimation accuracy influences performance and feasibility of the rehabilitation application.

\subsection{Limitations}

Four additional limitations should be taken into consideration when interpreting the findings of the current study. The first limitation is that participants interacted with only three objects, which can limit generalization. The number of objects is lower than recent work assessing HoloLens~2 joint prediction accuracy~\cite{bertolasi_evaluation_2025} and other egocentric datasets involving object interactions (e.g., H2O~\cite{kwon_h2o_2021}). However, the three objects were specifically chosen because they require different therapeutically and clinically relevant grasp types (i.e., cylindrical, lateral key pinch and tip to tip pinch). The second limitation is that the current dataset only included unimanual interactions (cf.~\cite{kwon_h2o_2021}). That is, bimanual tasks would create different occlusion scenarios that could not be investigated here. Further, restricting the movement to one hand may not be representative of how individuals with hand impairment would naturally interact with objects (e.g., using the left hand to aid the right hand in lifting the object). However, having participants perform unimanual tasks matches therapeutic and clinical assessments (e.g., GRASSP;~\cite{kalsi-ryan_graded_2012}). The third limitation is the method employed to generate the ground truth annotations (i.e., triangulation from a multi-camera setup). Although ground truth estimates were generated from validated pose estimation methods, the estimates may be less accurate than using marker-based systems which are considered gold-standard references. The current study employed a conservative outlier rejection procedure to minimize the impact of potentially inaccurate annotations. It is also important to consider the dynamic nature of the tasks employed in the current study (cf.~\cite{bertolasi_evaluation_2025}). That is, markers on each joint could have influenced pose estimation accuracy of the state-of-the-art algorithms or HoloLens~2 by causing additional occlusions, and could potentially limit the usefulness of the dataset. Lastly, the limited field of view of HMD egocentric camera made the task difficult for individuals with higher levels of hand impairment. That is, additional movements of the head and torso when trying to interact with objects made it difficult to keep the hands and calibration cube in the egocentric camera's field of view. This resulted in two participants being excluded from the analysis (i.e., average GRASSP Prehension Subscore = 7). Although this potentially limits our dataset to only including individuals with cSCI and relatively high hand function, the included participants have a full range of hand impairment (i.e., GRASSP Prehension Subscores range from 4 to the maximum score of 30).

\section{Conclusion}
\label{sec:conclusion}
Pose estimation accuracy during therapy-relevant tasks involving real-object interactions did not differ between a group of individuals with cSCI and uninjured controls. Further, the transparency of the object did not meaningfully influence pose estimation accuracy. While state-of-the-art pose estimation methods showed improved accuracy over the onboard HoloLens~2 hand tracking system, future work should investigate whether the modest advantages are meaningful and outweigh potential tradeoffs of integration onto HMDs. Overall, the findings suggest that the HoloLens~2 AR system can be explored when creating AR rehabilitation applications for individuals with hand impairment, and scientists can use the dataset to test and refine future pose estimations methods with a hand-impaired population.

\section{Data Availability}
\label{sec:data_availability}

3D ground truth estimates, 3D estimates from the HoloLens~2, and individual image frames that were included in the analyses will be made available upon journal publication. 

\section*{Acknowledgments}

The authors would also like to acknowledge Sharmini Atputharaj for leading participant recruitment, Seerat Choudhry for assistance processing data, and all the participants in the study.



\bibliographystyle{IEEEtran} 
\bibliography{IEEEabrv,SCIRehab}  

\end{document}